\documentclass[12pt,reqno]{amsart}

\usepackage{amsthm,amsmath,amsfonts,amssymb}
\usepackage{ascmac}
\usepackage{bm}
\usepackage[numbers]{natbib}
\usepackage{here,comment}
\usepackage{ascmac}
\usepackage{fancybox}
\usepackage{graphicx}
\usepackage{xcolor}
\usepackage{tabularx}
\usepackage{booktabs}
\usepackage{algorithm}
\usepackage{algorithmicx}
\usepackage{hyperref}

\usepackage{amsaddr}
\usepackage{fancyhdr}
\usepackage{natbib}
\usepackage{fullpage}
\usepackage{newtxtext,newtxmath}

\newtheorem{theorem}{Theorem}

\newtheorem{assumption}{Assumption}
\theoremstyle{definition}
\newtheorem{definition}{Definition}

\newtheorem{example}{Example}

\newcommand{\R}{\mathbb{R}}
\newcommand{\N}{\mathbb{N}}

\newcommand{\mB}{\mathcal{B}}

\newcommand{\mD}{\mathcal{D}}

\newcommand{\mP}{\mathcal{P}}

\newcommand{\mN}{\mathcal{N}}
\newcommand{\mX}{\mathcal{X}}

\newcommand{\Ep}{\mathbb{E}}
\renewcommand{\Pr}{\mathbb{P}}

\renewcommand{\hat}{\widehat}
\renewcommand{\tilde}{\widetilde}

\newcommand{\argmin}{\operatornamewithlimits{argmin}}
\newcommand{\argmax}{\operatornamewithlimits{argmax}}

\newcommand{\mone}{\textbf{1}}

\usepackage{mathtools}
\mathtoolsset{showonlyrefs=true}

\usepackage{amsaddr}

\title{Bayesian Inference for Consistent Predictions \\in Overparameterized Nonlinear Regression}

\author{Tomoya Wakayama}

\address{The University of Tokyo}

\date{\today, \textit{Contact}: \textit{tom-w9@g.ecc.u-tokyo.ac.jp}}

\allowdisplaybreaks

\begin{document}

\begin{abstract}
The remarkable generalization performance of large-scale models has been challenging the conventional wisdom of the statistical learning theory. Although recent theoretical studies have shed light on this behavior in linear models and nonlinear classifiers, a comprehensive understanding of overparameterization in nonlinear regression models is still lacking. This study explores the predictive properties of overparameterized nonlinear regression within the Bayesian framework, extending the methodology of the adaptive prior considering the intrinsic spectral structure of the data. Posterior contraction is established for generalized linear and single-neuron models with Lipschitz continuous activation functions, demonstrating the consistency in the predictions of the proposed approach. Moreover, the Bayesian framework enables uncertainty estimation of the predictions. The proposed method was validated via numerical simulations and a real data application, showing its ability to achieve accurate predictions and reliable uncertainty estimates. This work provides a theoretical understanding of the advantages of overparameterization and a principled Bayesian approach to large nonlinear models.
\end{abstract}

\maketitle

\section{Introduction}
In recent years, the field of machine learning has witnessed a paradigm shift in understanding generalization performance. Traditional statistical theory emphasizes the importance of controlling model complexity and avoids overfitting by balancing the bias-variance trade-off \citep{vapnik1999overview}. However, with the emergence of large-scale deep neural networks, which often have considerably more parameters than the number of training data samples, conventional wisdom has been challenged. These models have demonstrated the ability to achieve high accuracy on test data by completely interpolating the training data. This discrepancy between empirical observations and traditional learning theory has necessitated a new explanation \citep{zhang2021understanding}.

A groundbreaking study by \cite{bartlett2020benign} provides a theoretical foundation for overparameterized linear regression models through a non-asymptotic analysis of the generalization error of a minimum norm interpolator. This discovery has reshaped our understanding of generalization performance in large statistical models, suggesting that overparameterization does not necessarily lead to poor prediction, even in simple models. Instead, the key to good performance lies in the model's expressiveness and its ability to capture the essential structure of the data. Recent theoretical studies have demonstrated that overparameterization leads to accurate predictions in various statistical models, shedding light on the behavior of modern machine learning models.

While the theoretical understanding of the goodness of overparameterization has advanced significantly in linear estimation, such as in linear regression \citep{bartlett2020benign, tsigler2020benign} and kernel methods \citep{liang2020just, li2023benign}, the analysis of nonlinear estimators remains limited. Although some studies have investigated this issue in nonlinear neural networks for classification problems \citep{frei2022benign, kou2023benign}, a comprehensive understanding is lacking as to why and under what conditions overparameterized nonlinear regression can achieve good generalization performance. Hence, addressing this question is crucial for reinforcing the theoretical foundations of modern machine learning.

In this paper, the predictive properties of overparameterized nonlinear regression models within the Bayesian framework are explored. Bayesian inference combines the prior distribution of parameters with a likelihood function to obtain the posterior distribution. \cite{wakayama2023bayesian} proposed a methodology that incorporates the structure of the data into the prior distribution, capturing the linear model structure as the likelihood, which results in accurate predictions even in non-sparse high-dimensional models. The key idea is to adaptively adjust the prior variance based on the intrinsic structure (spectra) of the data, which is found to be critical in prediction error convergence in the sense of posterior contraction. We extend this to nonlinear regression models, specifically single-neuron models~\citep{pmlr-v125-yehudai20a,pmlr-v178-shamir22a} and generalized linear models (GLMs) \citep{mccullagh2019generalized} and theoretically analyze the prediction error of the resulting posterior distribution. The main contributions of this study are summarized as follows:
\begin{itemize}
    \item For generalized linear models based on one-parameter exponential families, including logistic and Poisson regressions, the conditions under which the posterior distribution converges to the true parameter are established in terms of predictive risk, providing an upper bound on the convergence rate (Theorem~\ref{thm:expfam}). This finding highlights the applicability of the proposed approach to discrete response-variable regression, such as those encountered in classification problems.
    \item The proposed method is demonstrated to generalize well to single-neuron models with a Lipschitz continuous nonlinear activation function and Gaussian noise (Theorem~\ref{thm:gauss}). This result suggests that even in more general models involving nonlinearity, overparameterization allows benign prediction under appropriate prior distributions.
    \item The proposed Bayesian approach not only provides a theoretical guarantee of the generalization performance of the overparametrized models but also estimates the uncertainty in the predictions. By sampling from the posterior distribution, a confidence level can be constructed for each prediction, which is essential in decision-making processes.
\end{itemize}

\subsection{Related work}
Bayesian GLM has been studied intensively for high dimensions. The first paper to study high-dimensional Bayesian GLM \citep{ghosal1997normal} realized posterior concentration for the parameter without assuming any sparsity structure by imposing a restricted growth condition on the parameter dimension. This underparameterized setting differs from the present work. Subsequently, several studies explored the GLM setting where the number of parameters may exceed the sample size \citep{jiang2007bayesian, liang2013bayesian, jeong2021posterior}. These works have demonstrated the asymptotic properties of the posterior distribution; however, they assume sparsity, implying that the parameters are inherently low-dimensional. The properties of Bayesian variable selection in GLMs have also been investigated in similar domains \citep{chen2008bayesian, cao2020variable, narisetty2018skinny, lee2021bayesian, jeong2022posterior}. However, the assumption of parameter sparsity can be stringent, and the current study is conducted in a setting that does not rely on this assumption, thus expanding the scope of research in this area. The methods and theory of Bayesian inference in high-dimensional models are summarized in \cite{banerjee2021bayesian}.

Numerous theoretical results suggest that overparameterization leads to good generalization performance. \cite{bartlett2020benign} demonstrated that the excess risk converges to zero in overparameterized linear models. This theory has been widely extended to other linear estimators, such as regularized regression~\citep{tsigler2020benign,koehler2021uniform,pmlr-v151-wang22k,li2021minimum,cheng2024dimension}, kernel regression~\citep{liang2020just,liang2020multiple}, linear neural networks~\citep{JMLR:v23:21-1011,suzuki2024optimal}, and regression for dependent data~\citep{nakakita2022benign,tsuda2023benign}. \cite{frei2022benign,cao2022benign,kou2023benign} ventured beyond these linear frameworks to investigate the benign nature of overparameterization, but their focus was limited to classification problems. \cite{pmlr-v178-shamir22a} tackled nonlinear regression and demonstrated the absence of benign overfitting, but their assumptions differ from the present work mainly in terms of distribution of the covariate.

\subsection{Organization}
The remainder of this paper is structured as follows. In Section~\ref{section:one-GLM}, the problem setting for one-parameter generalized linear models under the overparameterization assumption is introduced, providing a detailed description of the proposed methodology. Section~\ref{section:main} presents the theoretical results on posterior contraction for GLMs. Section~\ref{section:gauss} extends the results to the two-parameter case. Section~\ref{section:simulation} reports on the simulation studies conducted to validate the theoretical findings and assess the uncertainty estimation. Section~\ref{section:application} showcases a real data application, comparing the proposed approach with baseline methods in terms of classification performance.
Section~\ref{section:discussion} concludes the paper, discussing the contribution and limitations of this work. The proof and discussion of the main theorems and details of the computational algorithms used in the analysis are provided in the Appendix.

\subsection{Notation}
For $q\in \mathbb{N}$ and any vector $\bm{v}$, $\|\bm{v}\|_q$ denotes the $\ell_q$-norm of $\bm{v}$. 
For any vector $\bm{v}$ and symmetric matrix of the same dimension $A$, we define $\|\bm{v}\|_{A} = \sqrt{\bm{v}^{\top}A\bm{v}}$. 
For sequences $\{a_n\}_{n \in \N}$ and $\{b_n\}_{n \in \N}$, $a_n \gtrsim b_n$ denotes the existence of a constant $c > 0$ such that $a_n \geq c b_n$ for all $n \geq \overline{n}$ with some finite $\overline{n} \in \N$, and $a_n \lesssim b_n$ denotes the opposite; $a_{n}=o(b_{n})$ indicates that $a_{n}/b_{n}\to 0$ as $n \to \infty$, whereas $a_{n}=\omega(b_{n})$ indicates that $a_{n}/b_{n}\to\infty$ as $n \to \infty$; $a_n \gtrapprox b_n$ indicates that $a_n \gtrsim b_n (\log n)^c$ holds for every $c > 0$, and $a_n \lessapprox b_n$ denotes the opposite. For natural numbers $a,b \in \N$ with $a < b$, we define $[a:b] := \{a,a+1,a+2,...,b\}$. For any square matrix $A$, let $A^{-1}$ be the inverse matrix of $A$, $A^{\dagger}$ be the Moore--Penrose pseudoinverse matrix, and $\mathrm{tr}(A)$ be the sum of the diagonals of $A$. For a pseudo-metric space $(S,d)$ and a positive value $\delta$, $\mN(\delta, S,d)$ denotes the covering number, that is, the minimum number of balls that cover $S$ with a radius $\delta$ in terms of $d$. For two probability measures $\Pi_1$ and $\Pi_2$ on the same measurable space $(\mX,\mB)$, the total variation distance is given by $\|\Pi_1-\Pi_2\|_{\mathrm{TV}}:= \sup_{B\in\mB}| \Pi_1(B)-\Pi_2(B)|$, and the Kullback--Leibler (KL) divergence and the KL variation (also referred to as the centered second-order KL divergence) between them are respectively defined as $K(\Pi_1, \Pi_2)=\int_{\mX} \log(\mathrm{d}\Pi_1/\mathrm{d}\Pi_2) \mathrm{d}\Pi_1$ and $V(\Pi_1, \Pi_2)=\int_{\mX} \{\log(\mathrm{d}\Pi_1/\mathrm{d}\Pi_2)-K(\Pi_1, \Pi_2)\}^2\mathrm{d}\Pi_1$ if $\Pi_1$ is absolutely continuous with respect to $\Pi_2$; otherwise, both are $\infty.$ The Hellinger distance between two probability distributions $\Pi_1$ and $\Pi_2$ is defined as $H(\Pi_1, \Pi_2) = \sqrt{1 - \int_{\mX} \sqrt{\mathrm{d}\Pi_1 \mathrm{d}\Pi_2}}$.

\section{One-parameter Generalized Linear Regression Model} \label{section:one-GLM}
This section presents a comprehensive overview of the one-parameter generalized linear regression model, including the problem setting (\S\ref{subsection:setting}) and the composition of the predictor (\S\ref{subsection:method}). 

\subsection{Setting}\label{subsection:setting}
Consider the pair $(\bm{X},Y)$, where $\bm{X}$ is an $\R^p$-valued centered random covariate, and $Y$ is an $\R$-valued univariate response. The GLM is defined as follows:
\begin{equation}\label{eq:model}
\Ep[Y \mid \bm{X}=\bm{x}] = g(\bm{x}^{\top} \bm{\beta}^*) ,
\end{equation}
where $\bm{\beta}^*\in \R^p$ is the true unknown $p$-dimensional regression coefficient, and $g:\R\to \R$ is a known Lipschitz continuous function. In this setting, the response variable $Y$ follows a general distribution belonging to the one-parameter exponential family with a density function given by
\begin{equation}
L(y; \mu_{x,\bm{\beta}} ) = \exp\{ a(\mu_{x,\bm{\beta}})y + b(\mu_{x,\bm{\beta}}) +c(y) \}, \label{eq:model-expfam}
\end{equation}
where $\mu_{x,\bm{\beta}}\in \R$ is the parameter; $a:\R\to\R$ and $b:\R\to\R$ are known continuously differentiable functions, with $a$ having a nonzero derivative. The function $c:\R\to\R$ is chosen such that $L(y; \mu_{x,\bm{\beta}} )$ is a proper probability density.

This formulation encompasses several well-known models: Normal linear models: $Y\mid \bm{X}$ follows a $g(\bm{x}^{\top} \bm{\beta}^*)$-mean Gaussian distribution with fixed variance, and $g$ is the identity function; Logistic regression model: $Y\mid \bm{X}$ follows a Bernoulli distribution with parameter $g(\bm{x}^{\top} \bm{\beta}^*)$, and $g$ is the logistic function; Poisson regression model: $Y\mid \bm{X}$ follows a Poisson distribution with parameter $g(\bm{x}^{\top} \bm{\beta}^*)$, and $g$ is the exponential function. Note that the link function $g$ is not restricted to being canonical, thereby providing flexibility in modeling the relationship between the linear predictor and response variable.

This study focuses on a non-sparse overparameterization setting, where $n$ independent and identically distributed copies of $(\bm{X},Y)$ are generated, and their realizations $\mD:= \{(\bm{x}_1,y_1),\ldots,(\bm{x}_n,y_n)\}$ are observed. From these observations, a $p$-dimensional estimate $\hat{\bm{\beta}}$ is constructed for prediction, with $p$ being larger than the sample size $n$. In the asymptotic setting, where $n$ diverges, $p$ possesses a larger order of divergence, i.e., $p=\omega(n)$.

The primary objective of this study is to examine the generalization performance of a predictor trained on the observations $\mD$ in the overparameterized domain. We adopt as the predictor the mean $g(\bm{x}^{\top} \bm{\beta})$, with the estimated parameter $\hat{\bm{\beta}}$ plugged in. Then, the notion of excess risk is introduced to quantify the generalization performance, measuring the difference between the oracle predictor $g(\bm{X}^{\top}\bm{\beta}^*)$ and the predictor $g(\bm{X}^{\top} \hat{\bm{\beta}})$:
\begin{equation*}
\|  g(\bm{X}^{\top} \hat{\bm{\beta}}) - g(\bm{X}^{\top}\bm{\beta}^*)  \|_{P_{\bm{X}},2} .
\end{equation*}
where $\|\cdot\|_{P_{\bm{X}},2}$ denotes the $L_2(P_{\bm{X}})$ norm with respect to the distribution $P_{\bm{X}}$ of the covariates. As the later analysis shows, the convergence of the excess risk is related to the distributional convergence of the estimated parameter $\hat{\bm{\beta}}$.

\subsection{Method} \label{subsection:method}
Predictor construction follows a systematic approach: first, the data are divided into two sets; then, a prior distribution of $\bm{\beta}$ is developed based on the empirical spectral information of $\Sigma$ from one set; next, the likelihood is computed in the other set; finally, the posterior distribution of $\bm{\beta}$ is computed based on the prior and likelihood.

Initially, we divide the dataset $\mD$ into two equal parts, $\mD_1$ and $\mD_2$, each containing $n/2$ observations. Although the split ratio is arbitrary, it should not depend on the sample size $n$. The data splitting allows the use of $\mD_1$ for constructing the prior distribution and $\mD_2$ for evaluating the likelihood function, ensuring that the data used in these two steps are independent.

Next, we consider the empirical covariance matrix of the covariates $\bm{X}$, defined as:
\begin{equation}
\hat{\Sigma} := \frac{2}{n} \sum_{i=1}^{n/2}(\bm{x}_i - \bar{\bm{x}})(\bm{x}_i - \bar{\bm{x}})^{\top},
\end{equation}
where $\bar{\bm{x}}\in\R^p$ is the sample mean of the covariates from $\mD_1$. The factor $2/n$ ensures that $\hat{\Sigma}$ is an unbiased estimator of the population covariance matrix, whereby a spectral decomposition of $\hat{\Sigma}$ is performed:
\begin{equation}
\hat{\Sigma} =\sum_{j=1}^{p}\hat{\lambda}_j\hat{\bm{v}}_j\hat{\bm{v}}_j^{\top},
\end{equation}
where $\{\hat{\lambda}_j\}_{j=1}^p$ and $\{\hat{\bm{v}}_j\}_{j=1}^p$ are the eigenvalues and eigenvectors of $\hat{\Sigma}$, respectively, with $\hat{\lambda}_1 \geq \hat{\lambda}_2 \geq \cdots \hat{\lambda}_p \geq 0$. Using the empirical spectra, we define a low-rank approximation of $\hat{\Sigma}$ as:
\begin{equation}
\hat{\Sigma}_{1:k}:= \sum_{j=1}^{k}\hat{\lambda}_j\hat{\bm{v}}_j\hat{\bm{v}}_j^{\top}.
\end{equation}
This approximation captures the most significant $k$ eigenvalues and eigenvectors of $\hat{\Sigma}$, effectively reducing the dimensionality of the covariate space while retaining the most informative directions of variation. In choosing $k$ as a tuning parameter to manage the computational cost, it has to be fixed to an interval of $[L_{\kappa}:U_{\kappa}]$ so that it satisfies Assumption~\ref{ass:comb}.

We then employ the following prior distribution for $\bm{\beta}$.

\begin{definition}[Prior Distribution with Effective Spectra]
\begin{equation}\label{prior:beta}
\pi_{\mathfrak{b}} (\bm{\beta}\mid \hat{\Sigma},k) = \frac{\exp\big(- \bm{\beta}^{\top} \hat{\Sigma}_{1:k}^{\dagger} \bm{\beta} \big)\mone\{ \|\bm{\beta}\|_2 \le R\} }{ \int_{\mathbb{R}^p} \exp\big(- \bar{\bm{\beta}}^{\top} \hat{\Sigma}_{1:k}^{\dagger} \bar{\bm{\beta}}\big) \mone\{ \|\bar{\bm{\beta}}\|_2 \le R\} \mathrm{d}\bar{\bm{\beta}} },
\end{equation}
where $R > 0$ is a fixed constant that determines the radius of the truncation sphere.
\end{definition}
Although this prior distribution resembles a $p$-variate Gaussian distribution with mean $\bm{0}$ and covariance $\hat{\Sigma}_{1:k}$, there are some important distinctions. First, the distribution is truncated, with the indicator function restricting the domain to a centered sphere of radius $R$. This truncation is necessary because the low-rank approximation $\hat{\Sigma}_{1:k}$ is singular, and defining the density function without support restrictions would cause the integrals over $\R^p$ to diverge. Second, unlike \cite{narisetty2018skinny,jeong2021posterior,tang2023empirical}, we impose a non-sparse prior distribution, even though some parameters may be redundant. This choice is motivated by the desire to retain all potentially relevant covariates in the data and avoid the challenges associated with sparse priors, such as the need to specify sparsity-inducing hyperparameters. 

Next, we consider the posterior distribution, which combines the prior distribution and the likelihood function using Bayes' Theorem.
\begin{definition}[Posterior Distribution]
The probability density function of the posterior distribution is given by:
\begin{equation}
\pi(\bm{\beta}\mid \mD,k) = \frac{\pi_{\mathfrak{b}}(\bm{\beta}\mid \hat{\Sigma},k) {\prod_{(y_i, \bm{x}_i) \in \mD_2}L(y_i; g(\bm{x}_i^{\top}\bm{\beta}) )}}{ \int_{\R^p} \pi_{\mathfrak{b}}(\bm{\beta}\mid \hat{\Sigma},k) {\prod_{(y_i, \bm{x}_i) \in \mD_2}L(y_i; g(\bm{x}_i^{\top}\bm{\beta}) )} \mathrm{d} \bm{\beta} }, \label{eq:density_posterior}
\end{equation}

For any measurable set $B \subset \R^p$, the posterior probability of $\bm{\beta} \in B$ is given by:
\begin{equation}
\Pi \left( \{\bm{\beta} : \bm{\beta} \in B \} \mid \mD,k \right) = \int_B \pi(\bm{\beta}\mid \mD,k ) \mathrm{d} \bm{\beta}. \label{eq:definition_posterior}
\end{equation}
\end{definition}
The Bayesian framework allows the quantification of events of interest, such as the probability of predictions being close to the true values through probability measures defined in this manner. Moreover, as demonstrated later, the distributional information provided by the posterior is valuable in practice.

Note that the information about the response variable $y_i$ in $\mD_1$ is discarded since this subset is only used to compute the sample covariance matrix of the covariates. If excluding this information seems inefficient, the roles of $\mD_1$ and $\mD_2$ can be switched to construct a new estimator (posterior distribution), and the two estimators can be averaged in a manner similar to the sample splitting scheme~\citep{chernozhukov2018double}. The theoretical results presented next preserve this averaging.

\section{Main Results}\label{section:main}
In this section, the main theoretical results are presented. Initially, the key assumptions that underlie our analysis (\S\ref{subsection:ass}) are introduced. These assumptions concern the spectral properties of the covariance matrix and the concentration of the empirical covariance matrix around its population counterpart. Subsequently, the posterior contraction for the GLM is established (\S\ref{subsection:result}). The theoretical results characterize the posterior contraction, elucidating the intricate interplay between dimensionality, sample size, and eigenvalue decay of the covariance matrix.

\subsection{Assumptions}\label{subsection:ass}
The initial focus is on the boundedness of the operator norm of $\Sigma$ and concentration of the empirical covariance matrix $\hat{\Sigma}$ around $\Sigma$.
\begin{assumption}\label{ass:emp}
The operator norm $\|\Sigma\|_{\mathrm{op}}$ is bounded, and there exist sequences $\{\rho_n\}_{ n \in \N}$ and $\{\ell_n\}_{n \in \N}$ of non-negative reals with $\rho_n, \ell_n =o(1)$ as $n \to \infty$, such that for any sufficiently large $n \in \N$,
\begin{align}
\Pr \left( \frac{\| \hat{\Sigma} - \Sigma \|_{\mathrm{op}}}{\|\Sigma\|_{\mathrm{op}}}  \leq  \rho_n \right) \geq 1 - \ell_n. \label{eq:emp}
\end{align}
\end{assumption}
This assumption ensures that the empirical covariance matrix concentrates around the population covariance matrix. The sequence $\rho_n$ quantifies the rate of this concentration, whereas $\ell_n$ controls the probability of the concentration event. Notably, this assumption is mild and includes not only the sub-Gaussian distributions commonly imposed in high-dimensional statistics but also log-concave distributions, which encompass most unimodal parametric distributions \cite{walther2002detecting,walther2009inference} that allow for heavy-tailed distributions such as the Laplace distribution. In contrast, other related literature impose more restrictive conditions such as sub-Gaussianity \cite{bartlett2020benign} or strong log-concavity \cite{frei2022benign}. In a high-dimensional setting, where many covariates are incorporated, the mildness of the distributional assumption is particularly important, as it allows for a wider range of applicable scenarios.

The next assumption concerns the behavior of the eigenvalues of $\Sigma$.
\begin{assumption}\label{ass:comb}
Given a covariance matrix $\Sigma$ satisfying~\eqref{eq:emp}, with eigenvalues $\{\lambda_k\}_{k=1}^p$, sequence $\{\rho_n\}_{n \in \N}$ from Assumption~\ref{ass:emp}, and a sequence of bounds $(L_{\kappa}, U_{\kappa})$, there exists a sequence $\{\varepsilon_n\}_{n \in \N}$ of positive reals such that the following hold as $n \to \infty$:
\begin{align}
\mathrm{(i)} &\quad  \varepsilon_n\to 0,\quad n^{-1/2} \lesssim\rho_n \lesssim \varepsilon_n^2,\qquad \mathrm{(ii)} \quad
(\rho_n +\lambda_k)\lambda_k \lessapprox \frac{\varepsilon_n^2}{k},\quad \forall k \in [ L{\kappa}:U_{\kappa}],  \\
\mathrm{(iii)} & \quad \sum_{k\in[L_{\kappa}+1: p]}\lambda_k \lesssim \varepsilon_n^4, \qquad
\mathrm{(iv)} \quad  U_{\kappa} \lessapprox n\varepsilon^2  .
\end{align}
\end{assumption}
This assumption imposes certain regularity conditions on the eigenvalues of $\Sigma$. Intuitively, conditions (i) and (ii) control the rate at which the bulk of the eigenvalues (those indexed from $L_{\kappa}$ to $U_{\kappa}$) are allowed to decay moderately. Condition (iii) requires that the eigenvalues beyond the bulk (those indexed larger than $U_{\kappa}$) are sufficiently small, whereas condition (iv) constrains the size of $U_{\kappa}$ relative to sample size $n$ and rate $\varepsilon_n$. These conditions are crucial for deriving the posterior contraction rates in Theorem~\ref{thm:expfam}.

Finally, we assume a certain local Lipschitz condition of the link function $g$ in terms of the Kullback-Leibler (KL) divergence and variational form between the corresponding probability distributions.
\begin{assumption}\label{ass:KLKV}
For any pair $\bm{\beta},\bm{\beta}'\in \R^p$, the GLM with link function $g$ satisfies
\begin{align*}
K(P, P'), V(P, P') \lesssim  \| \bm{\beta} - \bm{\beta}' \|_{\Sigma}^2,
\end{align*}
where $P$ and $P'$ are the probability distributions corresponding to the densities $L(y;g( \bm{x}^{\top} \bm{\beta}))$ and $L(y;g( \bm{x}^{\top} \bm{\beta}'))$, respectively.
\end{assumption}
This assumption limits the amount of variation in a distribution when different parameters are used, to control the complexity of the model and is satisfied by several common GLMs, as illustrated in the following examples.

\begin{example}\label{example:KLKV}
\begin{enumerate}
\item For normal linear regression with known variance $\sigma^2$, we have
\begin{align*}
2K(P, P') = V(P, P') \le \| g(\bm{X}^{\top}\bm{\beta}) - g(\bm{X}^{\top}\bm{\beta}') \|_{P_{\bm{X}},2}^2 \lesssim \| \bm{\beta} - \bm{\beta}' \|_{\Sigma}^2,
\end{align*}
where the last inequality holds if the link function $g$ is Lipschitz continuous.
\item For binary regression with the logistic link, we have
\begin{align*}
K(P, P'), V(P, P')  &\le \| \bm{\beta} - \bm{\beta}' \|_{\Sigma}^2,
\end{align*}
which are both bounded by $\| \bm{\beta} - \bm{\beta}' \|_{\Sigma}$. In contrast, for the probit link, Assumption~\ref{ass:KLKV} does not hold in general.

\item For Poisson regression with a bounded and positive link function $g$, we have
\begin{align*}
K(P, P'), V(P, P')  &\lesssim \| \bm{\beta} - \bm{\beta}' \|_{\Sigma}^2.
\end{align*}
Although the exponential function, $g(r)=e^r$, is the canonical link function for Poisson regression, it is unbounded, and Assumption~\ref{ass:KLKV} does not hold.
\end{enumerate}
\end{example}

These examples demonstrate that Assumption~\ref{ass:KLKV} is a reasonable condition that holds for several widely used GLMs. 

\subsection{Posterior Contraction}\label{subsection:result}

We now present the main result, which establishes the posterior contraction for the GLM under the assumptions introduced in the previous subsection.
\begin{theorem} \label{thm:expfam}
Consider the GLM \eqref{eq:model-expfam} with covariate distribution satisfying Assumptions \ref{ass:emp}-\ref{ass:KLKV}, and the prior distribution on $\bm{\beta}$ is \eqref{prior:beta}. Then, for a sufficiently large constant $M$, the posterior distribution contracts around the true parameter $\bm{\beta}^*$ as
\begin{equation*}
\Pi\Big({\bm{\beta}:\| g( \bm{X}^{\top}\bm{\beta})-g( \bm{X}^{\top}\bm{\beta}^*)\|_{P_{\bm{X}},2}> M\varepsilon_n }\mid ,\mD\Big) \overset{P^*}{\longrightarrow} 0 \quad \mbox{as}\quad n\to \infty.
\end{equation*}
Furthermore, if the eigenvalues of $\Sigma$ satisfy the additional condition $(nk)^{-1} \gtrsim (\rho_n +\lambda_k)\lambda_k$ for all $k \in [L_{\kappa} : U_{\kappa}]$, then the contraction rate can be explicitly characterized as
\begin{equation}
\varepsilon_n =O\left( U_{\kappa} \lambda_{L_{\kappa} }^{1/2} (\lambda_{L_{\kappa} }+\rho_n)^{1/2}  \right) .
\end{equation}
\end{theorem}
This theorem shows that under conditions suitable for the covariance matrix $\Sigma$ and prior distribution, the posterior distribution of the predictor concentrates around the oracle predictor at a rate $\varepsilon_n$ with respect to the $L_2(P_{\bm{X}})$ distance. The rate $\varepsilon_n$ depends on the interplay between sample size $n$, concentration rate $\rho_n$ of the empirical covariance matrix, and eigenvalues of $\Sigma$ in the bulk region.

\subsubsection{Consistency as a distribution}\label{subsubsection:dist}
While Theorem~\ref{thm:expfam} derives the posterior consistency of the predictor with respect to the $L_2$ distance, it also holds for other commonly used distances between probability distributions, such as the total variation distance, the Hellinger distance, and the KL divergence in several cases, including a list of distributions in Example~\ref{example:KLKV}. The contraction in these divergences is particularly useful for assessing the quality of the posterior distribution in approximating the true data-generating distribution.

\subsubsection{Consistency of Point Estimators}\label{subsubsection:point}
The posterior contraction result in Theorem~\ref{thm:expfam} not only provides a theoretical justification for using the posterior distribution as a reliable approximation of the true data-generating distribution but also has important implications for point estimators in the usual sense of consistency of estimators. Two commonly used point estimators in Bayesian inference are the maximum a posteriori (MAP) estimator and the posterior mean.

The MAP estimator, denoted as $\hat{\bm{\beta}}_{\text{MAP}}$, is defined as the mode of the posterior distribution:
\begin{equation*}
    \hat{\bm{\beta}}_{\text{MAP}} = \argmax_{\bm{\beta} \in \mathbb{R}^p} \Pi(\bm{\beta} \mid \mD).
\end{equation*}
Under the same assumptions as in Theorem~\ref{thm:expfam}, the MAP estimator can be shown to be consistent with respect to the $L_2(P_{\bm{X}})$ norm:
\begin{equation*}
    \| g( \bm{X}^{\top}\hat{\bm{\beta}}_{\text{MAP}})-g( \bm{X}^{\top}\bm{\beta}^*)\|_{P_{\bm{X}},2} \overset{P^*}{\longrightarrow} 0,\quad \mbox{as}~n\to \infty.
\end{equation*}

Another widely used point estimator is the posterior mean, denoted as $\hat{\bm{\beta}}_{\text{PM}}$, which is defined as the expectation of the parameter with respect to the posterior distribution:
\begin{equation}
\hat{\bm{\beta}}_{\text{PM}} = \int \bm{\beta} d\Pi(\bm{\beta} \mid \mD) .
\end{equation}
Under the assumptions of Theorem~\ref{thm:expfam} and the additional condition that posterior consistency with respect to the total variation distance or Hellinger distance holds (as discussed in \S\ref{subsubsection:dist}), the posterior mean is also consistent:
\begin{equation}
\| g( \bm{X}^{\top}\hat{\bm{\beta}}_{\text{PM}})-g( \bm{X}^{\top}\bm{\beta}^*)\|_{P_{\bm{X}},2} \overset{P^*}{\longrightarrow} 0,\quad \mbox{as}~n\to \infty.
\end{equation}
As the posterior mean considers the weights of the entire distribution (defined by integral), its consistency requires the additional condition that the posterior distribution is not ill-behaved. Further discussion is deferred to the Appendix.

\section{Extension to Two-Parameter Gaussian Case}\label{section:gauss}
In this section, the analysis is extended to the setting of GLM with a Gaussian distribution for the measurement error. 
Specifically, we consider the single-neuron model~\citep{pmlr-v178-shamir22a,pmlr-v125-yehudai20a}:
\begin{equation} \label{eq:model-gauss}
Y = g(\bm{X}^{\top} \bm{\beta}) + \varepsilon ; \quad \varepsilon\sim N(0,\sigma^2),
\end{equation}
where $\sigma>0$ is the standard deviation of the error term. This model is more flexible than the standard linear regression model and can accommodate various link functions $g$. For instance, if $g$ is a censored link function such as the ReLU, tanh, or softplus, then \eqref{eq:model-gauss} represents a censored regression model.

Here, after specifying the prior distributions for the parameters $\bm{\beta}$ and $\sigma^2$, the prior distribution~\eqref{prior:beta} for $\bm{\beta}$ is used. Regarding the variance parameter, we employ as the prior distribution, the inverse Gaussian distribution $\Pi_{\varsigma^2}(\cdot)$ with density of the form:
\begin{equation}
\pi_{\varsigma^2}(\sigma^2) = \sqrt{\frac{\xi}{2 \pi \sigma^6}}\exp \left( \frac{- \xi (\sigma^2 - \eta)}{2 \eta^2 \sigma^2} \right), \label{prior:sigma}
\end{equation}
where $\eta>0$ and $\xi>0$ denote the mean and shape parameters, respectively, and the hyperparameters are arbitrarily assigned. An important property of the inverse Gaussian prior is that the tails on both sides decrease rapidly, which is critical for posterior contraction. In contrast, the inverse gamma distribution, the conjugate prior to the Gaussian likelihood, has a light lower tail and a heavy upper tail. From a Bayesian perspective, the inverse Gaussian prior reflects the analyst's belief that the parameter will not have extremely large or small values. This prior distribution for the variance parameter has also been employed in \cite{szabo2013empirical, ning2020bayesian, wakayama2023bayesian}.

We now present an analog of Theorem~\ref{thm:expfam} for the two-parameter Gaussian model \eqref{eq:model-gauss}. 
\begin{theorem} \label{thm:gauss}
Consider the regression model~\eqref{eq:model-gauss} with covariate distribution satisfying Assumptions~\ref{ass:emp}-\ref{ass:KLKV} and a prior distribution on $(\bm{\beta}, \sigma^2)$ satisfying \eqref{prior:beta} and \eqref{prior:sigma}. Then, for a sufficiently large constant $M$, the following convergences hold as $n \to \infty$:
\begin{align}
&\Pi\Big(\{\bm{\beta}:\| g( \bm{X}^{\top}\bm{\beta})-g( \bm{X}^{\top}\bm{\beta}^*)\|_{P_{\bm{X}},2}> M\varepsilon_n \}\mid \,\mD\Big) \overset{P^*}{\longrightarrow} 0,\\
&\Pi\Big(\{\sigma^2:|\sigma^2-(\sigma^*)^2|> M\varepsilon_n \} \,\mid \,\mD\Big) \overset{P^*}{\longrightarrow} 0,
\end{align}
where $(\bm{\beta}^*, \sigma^*)$ denotes the true parameter pair. Furthermore, if the eigenvalues of $\Sigma$ satisfy the additional condition $(nk)^{-1} \gtrsim (\rho_n +\lambda_k)\lambda_k$ for all $k \in [L_{\kappa} : U_{\kappa}]$, then the contraction rate can be explicitly characterized as
\begin{align}
    \varepsilon_n =O\left( U_{\kappa} \lambda_{L_{\kappa} }^{1/2} (\lambda_{L_{\kappa} }+\rho_n)^{1/2}  \right) .
\end{align}
\end{theorem}

This theorem provides the posterior contraction for both predictor $g(\bm{X}^{\top}\bm{\beta})$ and error variance $\sigma^2$ in the two-parameter Gaussian model. The contraction of the predictor is with respect to the $L_2(P_{\bm{X}})$ distance, while that of $\sigma^2$ is with respect to the absolute difference.

The extension of the two-parameter Gaussian model is particularly relevant to applications where the error variance is unknown and needs to be estimated from the data. By considering the joint posterior contraction rates for $(\bm{\beta}, \sigma^2)$, Theorem~\ref{thm:gauss} presents a theoretical justification for Bayesian inference in such models and offers guidance on the choice of prior distributions that can adapt to the unknown error variance.

\section{Simulation}\label{section:simulation}
This section confirms by simulation the statement of the theorem presented in \S\ref{section:main} and \S\ref{section:gauss} through simulations. The posterior predictive distribution of the proposed method is computed using variational inference, and the detailed algorithm is deferred to the Appendix.

\subsection{Logistic Regression}
Here, we consider a binary classification scenario where the response variable $Y_i$ takes values in $\{0, 1\}$. The data are generated according to the following model:
\begin{equation}\label{DGP:logostic}
Y_i = \mathrm{Ber}(g_{ls}(\bm{X}_i^{\top}\bm{\beta})); \quad i = 1, 2, ..., n,
\end{equation}
where $g_{ls}(r) := 1/(1+e^{-r})$ is the logistic function. Specifically, for $n=50,100,\ldots,500$, to ensure an overparameterized setting, $\bm{\beta}\sim N(\bm{0},I_p)$ are generated with $p=n^{4/3}$. Two settings are considered for the covariate: (A) Gaussian covariate, $\bm{X}_i\sim N(\bm{0},\Sigma)$; (B) Laplacian covariate, $\bm{X}_i\sim L(\bm{0},\Sigma)$, where the eigenvalues of $\Sigma$ follow $\lambda_j(\Sigma)=10\exp(-j/8) + n\exp\{-\sqrt{n}\}/p$. Then, each $Y_i$ corresponding to $X_i$ and $\beta$ is sampled from \eqref{DGP:logostic}.

For every $n$, $20$ different datasets were repeatedly generated through the above procedure and each was analyzed to determine the behavior of the proposed method in response to an increase in sample size. In each case, a test dataset of $1000$ instances was generated, to evaluate the performance of the trained classifier. The performance of the Bayesian predictions was evaluated, employing four widely used metrics: 0-1 loss, which quantifies the misclassification rate; area under the ROC curve (AUC), which assesses the discriminative ability of the classifier; rate of unconfident misclassifications (UM), which measures the proportion of unconfident predictions among misclassified instances; and confident correct (CC) predictions rate, which calculates the proportion of accurate predictions among confident ones. We refer to a confident prediction as one whose $95\%$ predictive interval does not cross the threshold $(0.5)$, whereas an unconfident one does. The latter two metrics are helpful when building systems that defer action when there is no confidence in the prediction. Additionally, principal component regression (PCR) \citep[][]{bishop2006pattern} was performed as a baseline method, with the number of principal components selected by 5-fold cross-validation from $\{1,2,\ldots,30\}$.

\begin{figure}[t]
\begin{center}
\includegraphics[width=\textwidth]{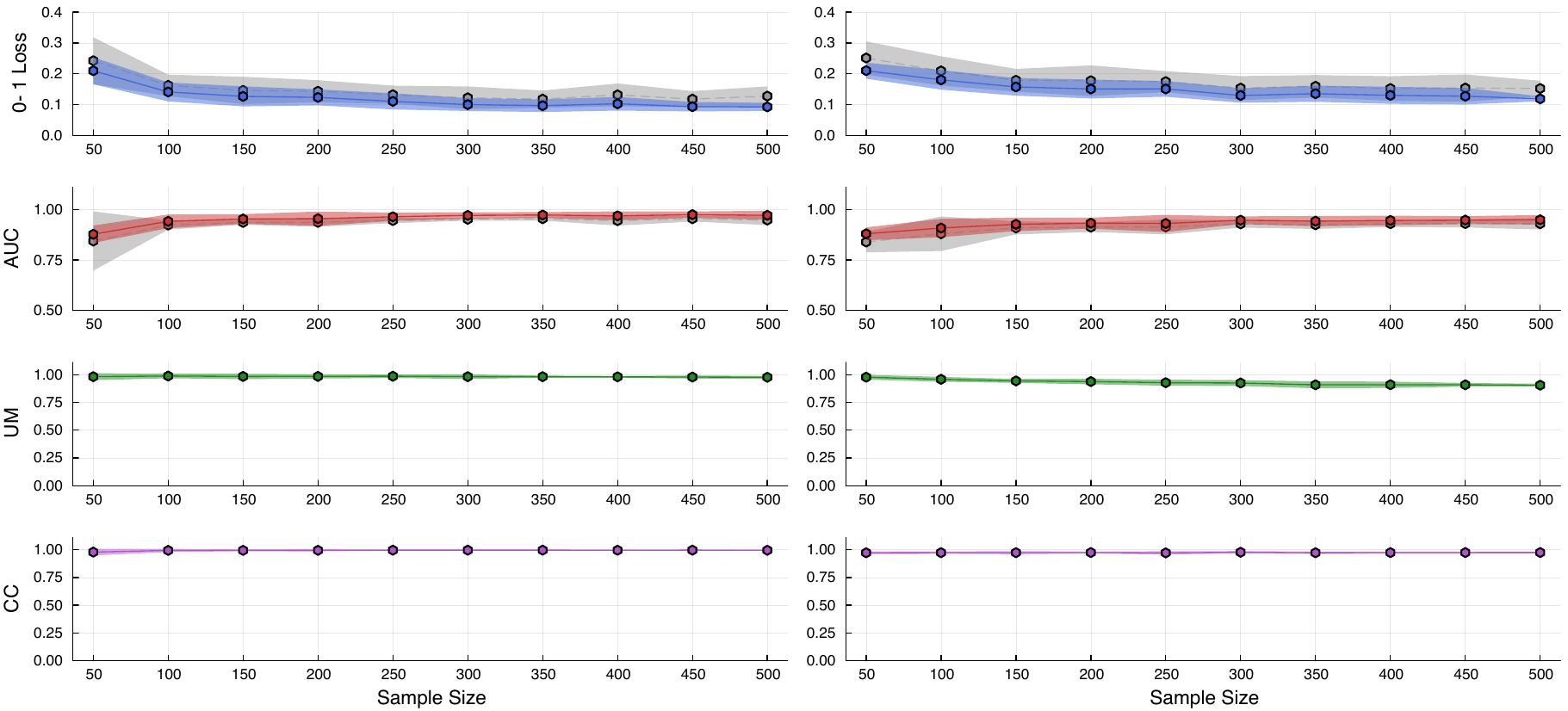}
\end{center}
\caption{Change of 0-1 loss, AUC (area under the ROC curve), UM (proportion of unconfident predictions among misclassifications), and CC (proportion of accurate predictions among confident ones) of the proposed method as sample size increases for the cases of logistic regression with Gaussian covariate (left) and Laplace covariate (right). The gray lines represent the results of the principal component regression. The shaded bands represent the 90\% interval.
 \label{fig:sim2}}
\end{figure}

Figure~\ref{fig:sim2} illustrates the results, where the reported values are averaged over the $20$ datasets. As sample size increases, the 0-1 loss exhibits a consistent decrease in both settings, indicating an improvement in classification accuracy. This trend persists even in the overparameterized setting, aligning with the theoretical findings. The AUC metric also shows a steady increase with sample size, suggesting an enhanced ability to discriminate between the two classes. Our method outperformed PCR, and these results underscore the importance of overparameterization for prediction. The UM and CC metrics provide insights into the uncertainty estimation of the proposed method. UM consistently presents a high value. This indicates that the method can warn us of a lack of confidence when misclassifying. The CC also presents a similarly high value, suggesting that confident predictions are trustworthy. These observations suggest that the proposed method effectively captures the uncertainty associated with predictions, thereby leveraging it to make more informed decisions.

\subsection{Nonlinear regression with Gaussian noise}
In this subsection, the scenario of true responses taking values along the positive real line $\R_+$ is considered. Specifically, the data are generated as follows:
\begin{equation}\label{DGP:nonlin}
Y_i=N(g_{sp}(\bm{X}_i^{\top}\bm{\beta}), 1);\quad  i=1,2,...,n,
\end{equation}
where $g_{sp}(r):=\log (1+e^r)$ is the softplus function. For $n=50, 100,\ldots,500$, we set $p = n^{4/3}$ and sample the coefficient $\bm{\beta}\in \R^p$ and covariate vectors $X_i\in\R^p$ in the same manner as in the previous experiment. Then, the scalar response is generated by \eqref{DGP:nonlin}.

For every $n$, $20$ distinct datasets were generated using the above procedure and, on each one, Bayesian prediction performance changes were examined with an increase in training data size. For a newly generated test dataset $\{(\bm{x}_i, g_{sp}(\bm{x}_i^{\top}\bm{\beta}))\}_{i=1}^{1000}$, the accuracy of Bayesian predictions was assessed using three widely adopted indicators: root-mean-squared error (RMSE), measuring the average deviation between predicted and true values; coverage probability (CP), evaluating the reliability of the $95\%$ prediction intervals by calculating the proportion of true values falling within the confidence intervals; and average length (AL), quantifying the width of the $95\%$ prediction intervals, providing insight into the precision of the predictions. Moreover, PCR was implemented, selecting the number of principal components using 5-fold cross-validation from $\{1,2,\ldots,30\}$.

\begin{figure}[t]
  \begin{center}
  \includegraphics[width=\textwidth]{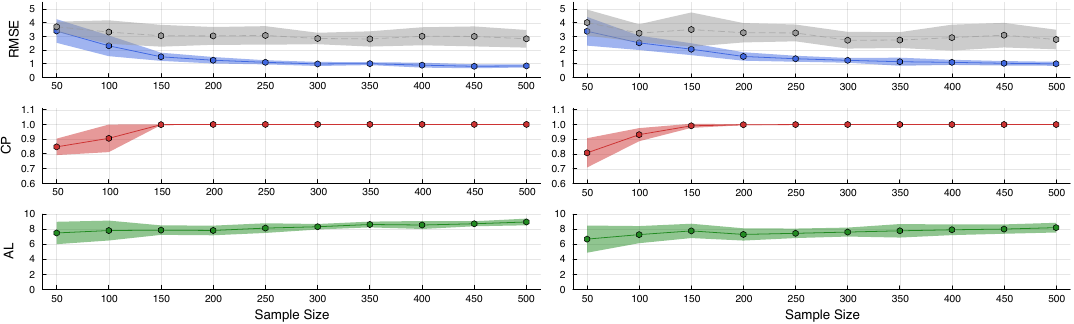}
  \end{center}
  \caption{Change of RMSE (root mean squared error), CP (coverage probability of $95$\% prediction interval), and AL (average length of $95$\% prediction interval) as sample size increases in nonlinear Gaussian regression with the Gaussian covariate case (left) and Laplace covariate case (right).\label{fig:sim1} }
\end{figure}

The results, averaged over $20$ datasets, are shown in Figure~\ref{fig:sim1}. First, the RMSE decreases monotonically as sample size increases. Even when $p$ is large and the setting is clearly overparameterized, the fact that accuracy continues to improve supports the theoretical results experimentally. Also, our method excels PCR, providing advantages of our method for prediction problems. Next, we focus on CP and AL, which are indicators related to prediction intervals, with CP representing the reliability of the prediction interval and AL representing its width. In the figure, the coverage is not sufficient when $n=100$; however, after that, CP stabilizes at a high value. AL also slightly increases but not by much, implying that the interval is not trivial. (Regardless of $n$, the mean of $\{g_{sp}(\bm{x}_i^{\top}\bm{\beta})\}_{i=1}^{1000}$ in the test data is around $3.5$, and the variance is around $5$.) These results suggest that the proposed method properly quantifies the uncertainty in constructing prediction intervals.

\section{Data Analysis} \label{section:application}
\subsection{Dataset}
We apply the proposed method to the ARCENE dataset, which was originally curated for the NIPS 2003 feature selection challenge for the binary classification problem. The dataset consists of mass spectrometric measurements from cancer patients and healthy individuals, with the goal of distinguishing between the two groups based on these measurements. The ARCENE dataset, accessible through the UC Irvine Machine Learning Repository~\citep{misc_arcene_167}, was constructed by merging three separate datasets: National Cancer Institute (NCI) ovarian cancer data, NCI prostate cancer data, and Eastern Virginia Medical School (EVMS) prostate cancer data. In total, the public dataset contains $200$ samples, comprising $88$ cancer samples and $112$ normal samples. These data are categorized as cancer and control samples, with $10000$ continuous covariates, with 7000 dimensions corresponding to real mass-spectrometry peaks and $3000$ dimensions representing random noise variables (probes). The primary challenges posed by the dataset lie in its high dimensionality and the presence of overlapping classes. The objective here is to discriminate between cancerous and normal samples.

\subsection{Experimental Setting}
The performance of the six distinct methods were compared.
\begin{itemize}
    \item Logistic Regression with effective spectral (ES) prior distribution: The proposed Bayesian methodology, as described in Section~\ref{subsection:method}, was applied to the logistic regression model to make predictions on unseen data. 
    
    \item Logistic Regression with the horseshoe (HS) prior: To handle the high-dimensionality of covariates, logistic regression was implemented using the Horseshoe prior~\citep{carvalho2009handling}. The "horseshoenlm" package in \texttt{R}~\citep{10.1111_rssc.12377} was used for the implementation.

    \item Logistic Regression with the normal (NM) prior: We investigated the performance of logistic regression with a standard normal prior distribution, which is the most commonly used prior and serves as a ridge-like regularization~\citep{gelman2013bayesian}. This prior provides flexibility in modeling the coefficient distribution and can be easily implemented using the Pyro probabilistic programming language~\citep{bingham2019pyro}.

    \item Logistic Regression with the Laplace (LP) prior: the performance of logistic regression was also explored using a standard Laplace prior distribution, which promotes sparsity in the coefficient estimates as a Bayesian Lasso~\citep{park2008bayesian}. This prior has proven effective in high-dimensional settings and can be efficiently implemented using Pyro, leveraging its support for non-Gaussian prior distributions~\citep{bingham2019pyro}.

    \item LightGBM: As a baseline method for table data prediction tasks, LightGBM~\citep{NIPS2017_6449f44a}, which is an efficient and scalable gradient boosting framework, was implemented adopting the "lightgbm" package in Python for the implementation and training of the model using Bayesian optimization~\citep{NIPS2012_05311655,frazier2018tutorial} via the "Optuna" package~\citep{optuna_2019}.
    
    \item Principal Component Regression: As a representative method for high-dimensional data, PCR was implemented with the number of bases selected from $1$ to $30$ by 5-fold cross-validation.
\end{itemize}

For all methods, performance evaluation was conducted using 5-fold cross-validation. In each fold, the model was trained on the training data, and predictions were made on the test data. As in the previous section, the evaluation metrics used to assess predictive performance were Accuracy, AUC, UM, and CC.

\subsection{Results}
The results of the experiment are presented in Table~\ref{tab:arecene}. The entries represent the average values of the metrics over $5$-fold cross-validation. Overall, among all the methods, logistic regression with the ES prior distribution achieves the highest accuracy and AUC, demonstrating its effectiveness in capturing the underlying structure of the high-dimensional ARCENE dataset. The superior performance of the ES prior can be attributed to its ability to adapt to the intrinsic dimensionality of the data by incorporating the empirical spectral information of the covariance matrix~\citep{wakayama2023bayesian}.

Let us take a closer look at the comparison. The horseshoe prior, known for its flexibility and sparsity-inducing properties~\citep{carvalho2009handling}, performs well, with the second-highest accuracy and AUC. This suggests that the regression on ARCENE dataset may exhibit some degree of sparsity, which the HS prior is able to capture effectively. However, the ES prior outperforms the HS prior, indicating that the adaptive nature of the ES prior is more beneficial in this context. Also, the fact that the $3000$ covariates are essentially irrelevant to the regression and have zero coefficients but are not highly sparse, may be the reason why the proposed method prevails. The LP and NM priors, while still providing competitive results, do not perform as well as the ES and HS priors. Nevertheless, their performance is still superior to the LightGBM baseline, highlighting the advantages of the Bayesian approach in handling excess covariates. PCR effectively selects relevant components from a large number of covariates although the results are only moderate and do not match those of our method which retains all covariates without reduction.

A notable aspect of the results is the uncertainty estimation of the Bayesian methods. The ES prior achieves the highest UM and CC values, indicating that it is highly effective in identifying uncertainty and providing reliable confidence estimates. This is particularly valuable in practical applications, where the ability to defer decision-making when the model is uncertain can prevent the risk of misclassifications. The HS prior also demonstrates good uncertainty quantification, with the second-highest UM value. However, the LP and NM priors exhibit very low UM values, suggesting that they may be overconfident in their predictions. Note that the LightGBM method does not provide uncertainty estimation, as it is not a probabilistic model.

\begin{table}[tb!]
    \centering
    \begin{tabular}{ccccc}
        \toprule
        {\bfseries Model} & {\bfseries Accuracy} & {\bfseries AUC} & {\bfseries UM} & {\bfseries CC}\\
        \midrule
        Logistic with ES & 0.895 & 0.950  & 0.860 & 0.963 \\
        Logistic with HS & 0.845 & 0.900 & 0.632 & 0.908 \\
        Logistic with LP & 0.805 & 0.886  & 0.009 & 0.936 \\
        Logistic with NM & 0.815 & 0.883  & 0.008 & 0.936 \\
        LightGBM   & 0.775 & 0.858 & -- & -- \\
        PCR & 0.810 & 0.878 & -- & -- \\
        \bottomrule
    \end{tabular}
    \caption{Comparison of classification performance among logistic regressions with effective spectral (ES, our method), horseshoe (HS), Laplace (LP), and normal (NM) priors, LightGBM, and principal component regression (PCR), using accuracy, area under the curve (AUC), proportion of uncertain predictions among misclassifications (UM), and proportion of correct predictions among certain classifications (CC) as metrics.  \label{tab:arecene} }
\end{table}

\section{Discussion} \label{section:discussion}
This study introduces a novel Bayesian approach for prediction in overparameterized generalized linear models, providing a theoretical foundation for practical applications. The primary theoretical contribution lies in establishing posterior contraction under appropriate conditions, validating the use of the posterior distribution as a reliable approximation of the true data-generating distribution. Empirical evidence from simulations and real-world datasets corroborates the effectiveness of the proposed method in delivering accurate predictions and reliable uncertainty estimates.

In the field of high-dimensional Bayesian inference, the proposed approach stands out by directly addressing non-sparsity while providing theoretical guarantees for prediction under nonlinearity. This unique perspective paves the way for Bayesian inference in overparameterized models.

This study marks a significant advance in overparameterization theory by achieving consistency in nonlinear regression problems. The robustness of posterior contraction theory to model complexity (nonlinearity) plays a pivotal role in attaining consistency, necessitating control of the covering number, decay of the prior distribution in the tail region, and concentration of the prior distribution around the true value of the parameter. 

Finally, the limitations of the paper can help identify potential areas for improvement. The theoretical analysis hinges on specific assumptions regarding the covariance matrix, which may not always hold in real-world scenarios. Future research can focus on relaxing these assumptions to extend the results to more general settings, as deep neural networks generalize to a wide range of data. Moreover, the computational complexity associated with sampling from the posterior distribution can pose challenges for large-scale datasets. Further improvements in uncertainty estimation can also be combined with more flexible approximation approaches, such as normalizing flow~\citep{JMLR:v22:19-1028,pmlr-v37-rezende15} and Stein variational gradient descent~\citep{NIPS2016_b3ba8f1b}.

\section*{Computer Programs}
The computer programs used in the paper for the numerical experiments in Section~\ref{section:simulation} and the application in Section~\ref{section:application} are publicly available on GitHub repository~\href{https://github.com/TomWaka/BA-Overparameterized-NonLinReg}{https://github.com/TomWaka/BA-Overparameterized-NonLinReg}. The computational algorithm is outlined in the appendix.

\section*{Acknowledgements}
I would like to express my sincere gratitude to my advisor, Masaaki Imaizumi, for his patient and rigorous supervision. I am also grateful to Shonosuke Sugasawa for his helpful suggestions. This work was supported by JSPS KAKENHI (22J21090) and JST ACT-X (JPMJAX23CS).

\appendix

\section{Proof of the Main Results}
In this section, proofs are provided for the main results presented in the previous sections. The key is to establish the posterior contraction rate for the squared $L_2(P_{\bm{X}})$ norm of the difference between the true and estimated mean functions, i.e., $\Ep [ \{ g(\bm{X}^{\top} \hat{\bm{\beta}}) - g(\bm{X}^{\top}\bm{\beta}^*) \}^2 ]$. Since the link function $g$ is assumed to be Lipschitz continuous, we have the following upper bound (up to a constant factor):
\begin{equation}\label{eq:L2_Sigma_norm}
    \|  g(\bm{X}^{\top} \hat{\bm{\beta}}) - g(\bm{X}^{\top}\bm{\beta}^*) \|_{P_{\bm{X}},2}  \lesssim \|\hat{\bm{\beta}} - \bm{\beta}^*\|_{\Sigma},
\end{equation}
where $\|\cdot\|_{\Sigma}$ denotes the norm induced by the covariance matrix $\Sigma$. This bound relates the contraction of the posterior distribution in the $L_2(P_{\bm{X}})$ norm to the contraction in the $\|\cdot\|_{\Sigma}$ norm, which is similar to the predictive risk in linear regression models \citep{bartlett2020benign, misiakiewicz2023six, cheng2024dimension}. Therefore, it suffices to show that the posterior distribution of $\|\hat{\bm{\beta}} - \bm{\beta}^*\|_{\Sigma}$ concentrates around zero at a certain rate.

To this end, we invoke a general result on posterior contraction rates from \cite{ghosal2000convergence}. Let $\mP = \mathbb{R}^p$ be the parameter space equipped with distance $d_\mP$ induced by the $\|\cdot\|_\Sigma$ norm, further defining it as a sequence of restricted parameter spaces
\begin{align}\label{eq:P_n}
\mP_n :=\{\bm{h}\in\mathbb{R}^p  : \|\bm{h}\|_2\leq H_n\},
\end{align}
where $H_n>0$ is a radius satisfying $H_n\to \infty$ as $n\to \infty$. Recall from Assumption \ref{ass:emp} that the empirical covariance matrix $\hat{\Sigma}$ concentrates around $\Sigma$ with probability at least $1-\ell_n$, which controls the randomness of the posterior distribution.

Theorem 2.1 of \cite{ghosal2000convergence} states that the posterior contraction in Theorem~\ref{thm:expfam} holds if there exists a sequence $H_n\to\infty$ such that the following three conditions are satisfied for any sufficiently large $n$ and some constant $C > 0$: 
\begin{align}
	&\log \mN(\varepsilon_n, \mP_n, d_\mP )\leq n\varepsilon_n^2,\label{ggv-thm2.1-1}
	\\
	&\Pi_{\mathfrak{b}}(\mP\setminus \mP_n\mid\hat{\Sigma})\leq \exp(-(C+4)n\varepsilon_n^2), \label{ggv-thm2.1-2}
	\\
	&\Pi_{\mathfrak{b}} \left(\left\{\bm{\beta}: K(P^*, P_{\bm{\beta}}) \leq \varepsilon_n^2, 
	V(P^*,P_{\bm{\beta}}) \leq \varepsilon_n^2 \right\}\mid\hat{\Sigma}\right)\geq \exp(-Cn \varepsilon_n^2).\label{ggv-thm2.1-3}
\end{align}
The first two conditions~\eqref{ggv-thm2.1-1} and~\eqref{ggv-thm2.1-2} are determined by the tuple $(\mP, \mP_n, d_\mP, \Pi_{\mathfrak{b}})$. Since these conditions have been verified in~\cite{wakayama2023bayesian} for the same choice of the parameter spaces and prior distribution, we omit their proofs here.

To establish the remaining condition~\eqref{ggv-thm2.1-3}, we observe that by Assumption~\ref{ass:KLKV},
\begin{align*}
  \Pi_{\mathfrak{b}} \left(\left\{\bm{\beta}: K(P^*, P_{\bm{\beta}}) \leq \varepsilon_n^2, 
	V(P^*,P_{\bm{\beta}}) \leq \varepsilon_n^2 \right\}\mid\hat{\Sigma}\right)
  \geq \Pi_{\mathfrak{b}} \left(\left\{\bm{\beta}: \|\bm{\beta} - \bm{\beta}^*\|_{\Sigma}^2 \leq \varepsilon_n^2 \right\}\mid\hat{\Sigma}\right).
\end{align*}
Therefore, it suffices to show that 
\begin{align}\label{eq:prior_mass}
  \Pi_{\mathfrak{b}} \left(\left\{\bm{\beta}: \|\bm{\beta} - \bm{\beta}^*\|_{\Sigma}^2 \leq \varepsilon_n^2 \right\}\mid\hat{\Sigma}\right)\geq \exp(-Cn \varepsilon_n^2).
\end{align}
The proof of~\eqref{eq:prior_mass} follows from a similar argument as in \cite{wakayama2023bayesian}.

The extension to the case where $\sigma^2$ is unknown, is straightforward. Taking $[1/n,n]$ as $\mP_n$ in the case of an inverse Gaussian distribution as a prior distribution for $\sigma^2$, the posterior contraction holds because $\mP_n$ dominates $\mP$ as $n$ increases and the remaining $\mP\setminus \mP_n$ part has only the negligible mass of the prior.

Combining the above arguments, we conclude that the conditions of Theorem 2.1 in \cite{ghosal2000convergence} are satisfied, and hence the posterior contraction in Theorem~\ref{thm:expfam} holds. The explicit characterization of the contraction rate $\varepsilon_n$ in terms of the eigenvalues of $\Sigma$ follows from a careful analysis of the interplay between the rate $\rho_n$ in Assumption~\ref{ass:emp}, upper and lower bounds $(L_{\kappa}, U_{\kappa})$ in Assumption~\ref{ass:comb}, and the additional assumption on the eigenvalue decay stated in Theorem~\ref{thm:expfam}.

\section{Consistency as a Distribution and Point Estimator}
Following the discussion on posterior distribution contraction, the convergence of distributions and estimators is further explored using various metrics.

To extend the discussion, let us revisit Example~1.
\begin{example} Recall that $K(P, P')$ is the KL divergence; $V(P, P')$ is the KL variation; and $H(P,P')$ is the Hellinger distance between models with different parameters.
\begin{enumerate}
\item For normal linear regression with known variance $\sigma^2$, we have
\begin{align*}
2K(P, P') = V(P, P') = H(P,P') \le \| \bm{\beta} - \bm{\beta}' \|_{\Sigma}^2.
\end{align*}

\item For binary regression with a logistic link, we have
\begin{align*}
K(P, P'), V(P, P'), H(P, P') \lesssim \| \bm{\beta} - \bm{\beta}' \|_{\Sigma}^2.
\end{align*}

\item For Poisson regression with a link function $g$, we have
\begin{align*}
K(P, P') & \lesssim C_K \| \bm{\beta} - \bm{\beta}' \|_{\Sigma}^2 ,\\
V(P, P') & \lesssim C_V \| \bm{\beta} - \bm{\beta}' \|_{\Sigma}^2 ,\\
H(P, P') & \lesssim C_H \| \bm{\beta} - \bm{\beta}' \|_{\Sigma}^2 ,
\end{align*}
where $C_K = C_{\bm{\beta}} \| g' / g\|_{\infty}^2 + C_{\bm{\beta}} \| g'' / g\|_{\infty} + \| g'' \|_{\infty}$, $C_V = C_{\bm{\beta}} \| g' / g\|_{\infty}^2$, $C_H = \| g' / \sqrt{g} \|_{\infty}$ and $C_{\bm{\beta}} = \sup_{\bm{x}} | g( \bm{\beta}^{\top} \bm{x} )|$. Note that the factors, $C_K, C_V,$ and $C_H$, are not necessarily bounded.
\end{enumerate}
\end{example}

As shown in the previous section, the consistency of $\| \bm{\beta} - \bm{\beta}' \|_{\Sigma}^2$ holds. Consequently, in the above examples, if the distance between distributions can be bounded by a constant times the quantity, the consistency of the posterior distribution to the true data-generating distribution is obtained with respect to the KL divergence, KL variation, and Hellinger distance. Furthermore, because the total variation and Hellinger distances induce the same topology, posterior consistency also holds in terms of the total variation distance.

In connection with distributional convergence, we now shift our focus to the convergence of the posterior mean. If the parameter space $\mP$ is convex and distance $d_{\mP}$ on $\mP$ is bounded and convex, then the posterior mean serves as a consistent estimator (\citep[e.g., Theorem 6.8 in][]{ghosal2017fundamentals}). The assumption of $\mP$ being convex is often satisfied, and the boundedness and convexity of the distance hold for the total variation distance. Hence, in the above examples, the consistency of the posterior mean is achieved.

\section{Computational Method}\label{sec:computation}
In this section, the computational method for the proposed approach is described. Stochastic variational inference~\citep{graves2011practical,hoffman2013stochastic,blundell2015weight} is employed to approximate the posterior distribution of the model parameters. Since the closed-form expression of the KL divergence is not available, a reparameterization trick~\citep{Kingma2014AutoEncodingVB,rezende2014stochastic} was used to enable efficient gradient-based optimization. The reparameterization trick allows the variational distribution to be expressed in terms of a deterministic function of the variational parameters and a random variable, making the gradients of the variational parameters computable. The optimization part was performed using the Adam optimizer~\citep{kingma2014adam}.

\subsection{Logistic Regression Example}
To illustrate the computational procedure, let us consider an example of logistic regression. The model components are as follows:
\begin{enumerate}
    \item Likelihood: $\prod_{(y_i, \bm{x}_i) \in \mD_2} L(y_i;\bm{x}_i^{\top}\bm{\beta} )$
    \item Prior: $p_{pri}=\phi(\bm{\beta}\mid \bm{0},\hat{\Sigma}_{1:k} )$
    \item Posterior: $p_{pos}\propto \phi(\bm{\beta}\mid \bm{0},\hat{\Sigma}_{1:k} )\prod_{(y_i, \bm{x}_i) \in \mD_2}\phi(y_i;\bm{x}_i^{\top}\bm{\beta},\sigma^2 )$
    \item Approximator: $q(\bm{\beta}\mid {\tilde{\bm{\mu}}},\tilde{\Sigma} )$
\end{enumerate}
Indeed, the prior distribution is not just a normal distribution but has a support constraint. However, it can be approximated as a normal distribution if $R$ is taken large enough. Also, a procedure that computes the posterior distribution without support constraints and then projects it onto the constrained region, was justified by~\citep{sen2022constrained}.

Our objectove is to find the optimal variational parameters ${\tilde{\bm{\mu}}}_{opt}$ and $\tilde{\Sigma}_{opt}$ that would minimize the KL divergence between the variational approximation $q(\bm{\beta}\mid {\tilde{\bm{\mu}}},\tilde{\Sigma} )$ and the true posterior $p_{pos}$:
\begin{equation}
    {\tilde{\bm{\mu}}}_{opt},\tilde{\Sigma}_{opt} := \argmin_{{\tilde{\bm{\mu}}},\tilde{\Sigma}} \mathrm{KL}(q(\bm{\beta}\mid {\tilde{\bm{\mu}}},\tilde{\Sigma} ) \|  p_{pos}),
\end{equation}
where $\mathrm{KL}(p \|q)$ denotes the KL divergence between two probability measures with densities $p$ and $q$.
The KL divergence can be expressed as
\begin{align}
    \mathrm{KL}(q(\bm{\beta}\mid {\tilde{\bm{\mu}}},\tilde{\Sigma} ) \|  p_{pos})  = \mathbb{E}_q[\log q] - \mathbb{E}_q[\log \phi(\bm{\beta}\mid \bm{0},\hat{\Sigma}_{1:k} )] - \sum_{i=1}^n \mathbb{E}_q [ \log L(y_i;\bm{x}_i^{\top}\bm{\beta} ) ] + \mathrm{const}.
\end{align}
To compute the gradients of the KL divergence with respect to the variational parameters, we employ a reparameterization trick. Specifically, denoting by $\tilde{\bm{\beta}}$ a sample from the variational distribution $q$, we parameterize it as $\tilde{\bm{\beta}} = \tilde{\bm{\mu}} + \tilde{\Sigma}_{1/2} \tilde{\bm{\epsilon}}$, where $\tilde{\bm{\epsilon}}$ is a sample from a standard Gaussian distribution $\mathcal{N}(\bm{0}; \bm{I}_p)$ and $\tilde{\Sigma}_{1/2}$ are obtained from a decomposition of $\tilde{\Sigma}$, i.e., $\tilde{\Sigma} = \tilde{\Sigma}_{1/2} \tilde{\Sigma}_{1/2}^{\top}$. With this reparameterization, the KL divergence can be approximated as an unbiased estimate
\begin{align}
    \mathrm{KL}(q(\bm{\beta}\mid {\tilde{\bm{\mu}}},\tilde{\Sigma} ) \|  p_{pos}) \approx \log q(\tilde{\bm{\beta}} \mid {\tilde{\bm{\mu}}},\tilde{\Sigma} ) - \log \phi(\tilde{\bm{\beta}}\mid \bm{0},\hat{\Sigma}_{1:k} ) - \sum_{i=1}^n  \log L(y_i;\bm{x}_i^{\top}\tilde{\bm{\beta}} )  + \mathrm{const}.
\end{align}

Now, the gradients of the KL divergence can be computed with respect to the variational parameters. The gradient with respect to $\tilde{\bm{\mu}}$ is given by
\begin{align}
    &\frac{\partial}{\partial \tilde{\bm{\mu}}} \mathrm{KL}(q(\tilde{\bm{\beta}}\mid {\tilde{\bm{\mu}}},\tilde{\Sigma} ) \|  p_{pos}) \\
    &= \frac{\partial}{\partial \tilde{\bm{\mu}}} \log q(\tilde{\bm{\beta}} \mid {\tilde{\bm{\mu}}},\tilde{\Sigma} ) - \frac{\partial}{\partial \tilde{\bm{\mu}}}\log \phi(\tilde{\bm{\beta}}\mid \bm{0},\hat{\Sigma}_{1:k} ) - \sum_{i=1}^n \frac{\partial}{\partial \tilde{\bm{\mu}}} \log L(y_i;\bm{x}_i^{\top}\tilde{\bm{\beta}} ) \\
    &= \bm{0} - \frac{\partial}{\partial \tilde{\bm{\mu}}} \frac{-(\tilde{\bm{\mu}} + \tilde{\Sigma}_{1/2} \tilde{\bm{\epsilon}})^{\top} \hat{\Sigma}_{1:k}^{\dagger} (\tilde{\bm{\mu}} + \tilde{\Sigma}_{1/2} \tilde{\bm{\epsilon}})}{2} - \sum_{i=1}^n \frac{\partial \bm{x}_i^{\top}\tilde{\bm{\beta}}}{\partial \tilde{\bm{\mu}}}\frac{\partial}{\partial \bm{x}_i^{\top}\tilde{\bm{\beta}}} \log L(y_i;\bm{x}_i^{\top}\tilde{\bm{\beta}} ) \\ 
    &= \hat{\Sigma}_{1:k}^{\dagger} (\tilde{\bm{\mu}} + \tilde{\Sigma}_{1/2} \tilde{\bm{\epsilon}}) - \sum_{i=1}^n ( y_i - g(\bm{x}_i^{\top}\tilde{\bm{\beta}}))  \bm{x}_i \\
    &= \hat{\Sigma}_{1:k}^{\dagger} \tilde{\bm{\beta}} - \sum_{i=1}^n ( y_i - g(\bm{x}_i^{\top}\tilde{\bm{\beta}}))  \bm{x}_i,
\end{align}
where $g(\cdot)$ is the logistic function. Similarly, the gradient with respect to $\tilde{\Sigma}_{1/2}$ is given by
\begin{align}
    &\frac{\partial}{\partial \tilde{\Sigma}_{1/2} } \mathrm{KL}(q(\tilde{\bm{\beta}}\mid {\tilde{\bm{\mu}}},\tilde{\Sigma} ) \|  p_{pos}) \\
    &= \frac{\partial}{\partial \tilde{\Sigma}_{1/2}} \log q(\tilde{\bm{\beta}} \mid {\tilde{\bm{\mu}}},\tilde{\Sigma} ) - \frac{\partial}{\partial \tilde{\Sigma}_{1/2}}\log \phi(\tilde{\bm{\beta}}\mid \bm{0},\hat{\Sigma}_{1:k} ) - \sum_{i=1}^n \frac{\partial}{\partial \tilde{\Sigma}_{1/2}} \log L(y_i;\bm{x}_i^{\top}\tilde{\bm{\beta}} ) \\
    &= \frac{\partial}{\partial \tilde{\Sigma}_{1/2}}  \frac{-\log(| \tilde{\Sigma}_{1/2} \tilde{\Sigma}_{1/2}^{\top}|)  - (\tilde{\Sigma}_{1/2} \tilde{\bm{\varepsilon}})^{\top}\tilde{\Sigma}^{\dagger} (\tilde{\Sigma}_{1/2} \tilde{\bm{\varepsilon}}) }{2} 
    -  \frac{\partial}{\partial \tilde{\Sigma}_{1/2}}  \frac{ - \tilde{\bm{\beta}}^{\top} \hat{\Sigma}_{1:k}^{\dagger} \tilde{\bm{\beta}} }{2} \\
    &\quad - \sum_{i=1}^n \frac{\partial \bm{x}_i^{\top}\tilde{\bm{\beta}} }{\partial \tilde{\Sigma}_{1/2}} \frac{\partial }{\partial \bm{x}_i^{\top}\tilde{\bm{\beta}}}\log L(y_i;\bm{x}_i^{\top}\tilde{\bm{\beta}} ) \\
    &=   - \tilde{\Sigma}_{1/2}^{\dagger \top} 
    + \frac{\partial}{\partial \tilde{\Sigma}_{1/2}} \frac{ (\tilde{\bm{\mu}} + \tilde{\Sigma}_{1/2} \tilde{\bm{\epsilon}})^{\top} \hat{\Sigma}_{1:k}^{\dagger} (\tilde{\bm{\mu}} + \tilde{\Sigma}_{1/2} \tilde{\bm{\epsilon}}) }{2} - \sum_{i=1}^n \frac{\partial \bm{x}_i^{\top}\tilde{\bm{\beta}} }{\partial \tilde{\Sigma}_{1/2}} \frac{\partial }{\partial \bm{x}_i^{\top}\tilde{\bm{\beta}}}\log L(y_i;\bm{x}_i^{\top}\tilde{\bm{\beta}} ) \\
    &=   - \tilde{\Sigma}_{1/2}^{\dagger \top} 
    + \frac{\partial}{\partial \tilde{\Sigma}_{1/2}} \frac{ (\tilde{\Sigma}_{1/2} \tilde{\bm{\epsilon}})^{\top} \hat{\Sigma}_{1:k}^{\dagger}  \tilde{\Sigma}_{1/2} \tilde{\bm{\epsilon}} }{2} 
    + \frac{\partial}{\partial \tilde{\Sigma}_{1/2}} \frac{ 2\tilde{\bm{\mu}}^{\top} \hat{\Sigma}_{1:k}^{\dagger}  \tilde{\Sigma}_{1/2} \tilde{\bm{\epsilon}} }{2} \\
    &\quad - \sum_{i=1}^n \frac{\partial \bm{x}_i^{\top}\tilde{\bm{\beta}} }{\partial \tilde{\Sigma}_{1/2}} \frac{\partial }{\partial \bm{x}_i^{\top}\tilde{\bm{\beta}}}\log L(y_i;\bm{x}_i^{\top}\tilde{\bm{\beta}} )  \\
    &=   - \tilde{\Sigma}_{1/2}^{\dagger \top} 
    + \hat{\Sigma}_{1:k}^{\dagger}\tilde{\Sigma}_{1/2} \tilde{\bm{\epsilon}} \tilde{\bm{\epsilon}}^{\top}
    + \hat{\Sigma}_{1:k}^{\dagger} \tilde{\bm{\mu}} \tilde{\bm{\epsilon}}^{\top} 
    - \sum_{i=1}^n  \bm{x}_i \tilde{\bm{\epsilon}}^{\top} (y_i - g(\bm{x}_i^{\top}\tilde{\bm{\beta}})) \\
    &=   - \tilde{\Sigma}_{1/2}^{\dagger \top} 
    + \hat{\Sigma}_{1:k}^{\dagger} \tilde{\bm{\beta}} \tilde{\bm{\epsilon}}^{\top} 
    - \sum_{i=1}^n  \bm{x}_i \tilde{\bm{\epsilon}}^{\top} (y_i - g(\bm{x}_i^{\top}\tilde{\bm{\beta}})),
\end{align}
where $\tilde{\Sigma}_{1/2}^{\dagger}$ denotes the pseudoinverse of $\tilde{\Sigma}_{1/2}$.

The derived gradients can be used in conjunction with the Adam optimizer to update the variational parameters $\tilde{\bm{\mu}}$ and $\tilde{\Sigma}_{1/2}$. The optimization procedure iteratively updates the variational parameters until the maximum number of iterations or convergence is reached.

\bibliographystyle{alpha}
\bibliography{main}

\end{document}